\documentclass[letterpaper, 10 pt, conference]{ieeeconf}  

\IEEEoverridecommandlockouts                              

\usepackage{graphics} 
\usepackage{epsfig} 
\usepackage{mathptmx} 
\usepackage{times} 
\usepackage{amsmath} 
\usepackage{amssymb}  
\usepackage{amssymb}  
\usepackage{color}
\usepackage{graphicx}
\usepackage[utf8]{inputenc}
\usepackage[table]{xcolor}
\usepackage{hyperref}

\title{\LARGE \bf
A 4-DoF Parallel Origami Haptic Device for \\Normal, Shear, and Torsion Feedback
}

\author{Sophia R. Williams$^{1}$, Jacob M. Suchoski$^{2}$, Zonghe Chua$^{3}$, and Allison M. Okamura$^{3}$
\thanks{*This work was supported in part by National Science Foundation grants 1812966 and 1830163, a Stanford Robotics Center Fellowship sponsored by FANUC, the Wu Tsai Neurosciences Institute at Stanford University, and a Stanford Bio-X Fellowship.}
\thanks{$^{1}$S.~R.~Williams is with the Electrical Engineering Department, Stanford University, CA 94305, USA. 
        {\tt\small sophiarw@stanford.edu}}%
        \thanks{$^{2}$J. M. Suchoski is with Intuitive Surgical, Sunnyvale, CA 94305, USA.
        {\tt\small jacob.suchoski@intusurg.com}}%
\thanks{$^{2}$A.~M.~Okamura and Z. Chua are with the Mechanical 
Engineering Department,  Stanford University, CA 94305, USA.
        {\tt\small aokamura@stanford.edu}, {\tt\small chuazh@stanford.edu}}%
}

\begin{document}

\maketitle
\thispagestyle{empty}
\pagestyle{empty}

\begin{abstract}

We present a mesoscale finger-mounted 4-degree-of-freedom (DoF) haptic device that is created using origami fabrication techniques. The 4-DoF device is a parallel kinematic mechanism capable of delivering normal, shear, and torsional haptic feedback to the fingertip. Traditional methods of robot fabrication are not well suited for designing small robotic devices because it is challenging and expensive to manufacture small, low-friction joints. Our device uses origami manufacturing principles to reduce complexity and the device footprint. We characterize the bandwidth, workspace, and force output of the device. The capabilities of the torsion-DoF are demonstrated in a virtual reality scenario. Our results show that the device can deliver haptic feedback in 4-DoFs with an effective operational workspace of 0.64\,cm$^3$ with $\mathbf{\pm 30 ^ \circ}$ rotation at every location. The maximum forces and torques the device can apply in the x-, y-, z-, and $\mathbf{\theta}$-directions, are $\pm$1.5\,N, $\pm$1.5\,N, 2\,N, and 5\,N$\mathbf{\cdot}$mm, respectively, and the device has an operating bandwidth of 9\,Hz.
\end{abstract}

\section{INTRODUCTION}

Haptics, the sense of touch, can improve task performance and realism \cite{khurshid2016effects, krogmeier2019human}. However, a bottleneck to making haptic devices ubiquitous consumer products is a lack of high-fidelity devices that are also simple and affordable. Many industrial haptic devices are desktop-mounted, or ``grounded'', limiting the movement of the user \cite{cini2005novel}. Researchers have begun investigating affordable wearable cutaneous devices for the fingertips, which allow users to move unconstrained in their environment \cite{pacchierotti2017wearable}.

\begin{figure}[]
\includegraphics[width=\columnwidth]{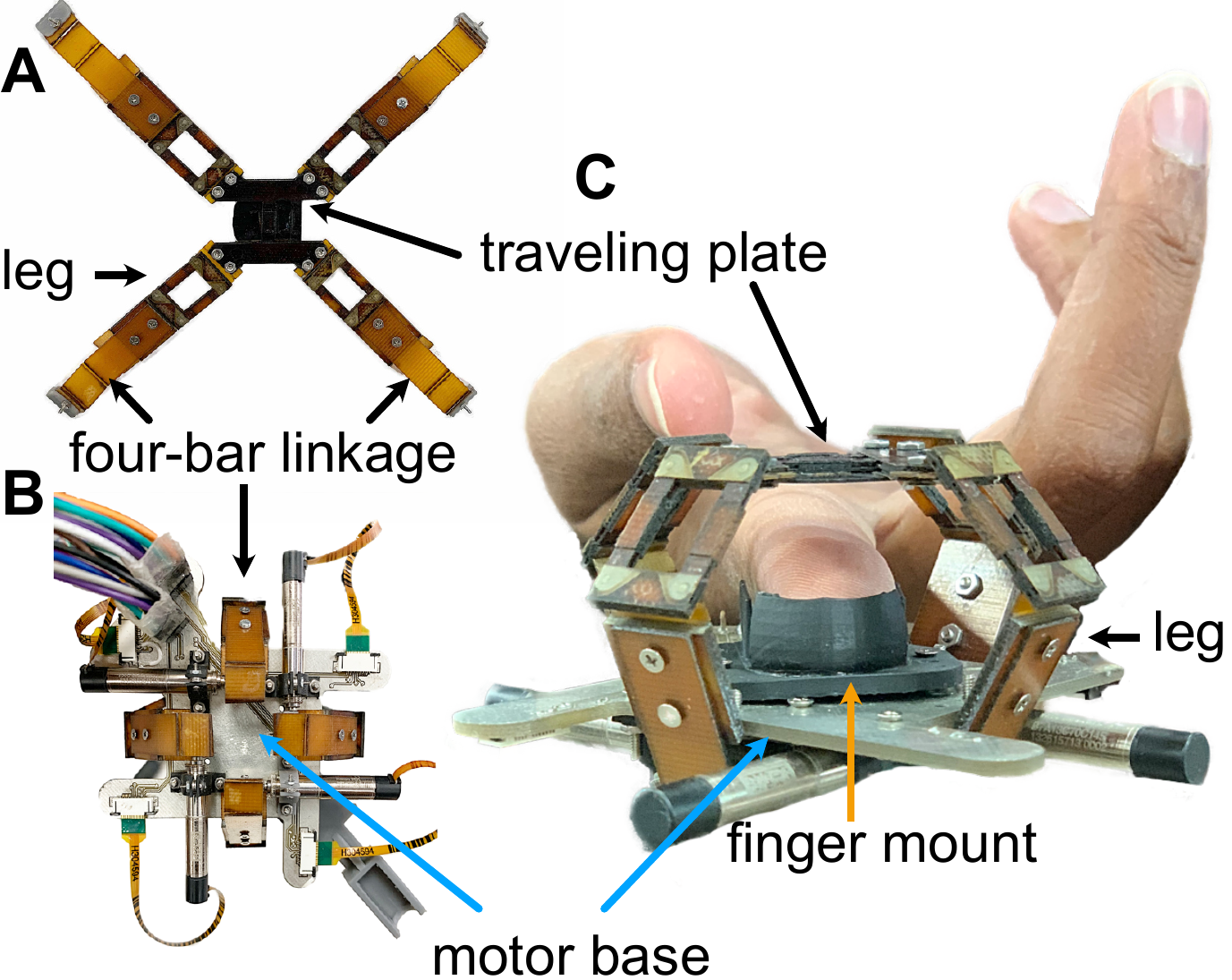}
    \caption{Different views of the 4-DoF origami cutaneous haptic device. (A) The origami device lays flat with the tactor, the device component that contacts the skin of the fingerpad, visible. (B) The motor base where the motors are mounted. Four bar linkages are attached to each motor shaft using a 3D printed interface. (C) The complete 4-DoF origami cutaneous haptic device. The device is mounted to the index finger of a user. The tactor is mounted to the traveling plate. The end effector can apply normal, x- and y-shear forces, and torsion haptic feedback. The device is a parallel mechanism with 4 legs. Each leg is actuated by its own motor that is connected to the mechanism using a four-bar linkage.}
    \label{fig:origami_device}
\end{figure}

Some fingertip mounted devices are small and stimulate few degrees-of-freedom (DoF), such as a 2-DoF wearable cutaneous device designed by Girard et al. that provides shear feedback in two directions \cite{girard2016haptip}. Futhermore, Schorr et al. developed a 3-DoF actuator inspired by the Delta mechanism that mounts to the fingertip and delivers shear and normal forces \cite{schorr2017three}. Similarly, Leonardis et al. designed and fabricated a 3-DoF device that uses both rotational and spherical DoF \cite{leonardisRSR}. Researchers have also developed 6-DoF wearable cutaneous devices but they are much bulkier than devices with fewer DoF \cite{young2019implementation, quek2015sensory}. In the current literature of wearable fingertip devices, it is apparent that as the a wearable cutaneous device's DoF increase, the device's size also increases. 

Haptic interaction in the real world, such as object manipulation, is composed of both kinesthetic and cutaneous feedback. Wearable fingertip devices move the grounding forces closer to area where the device is mounted, reducing the required size and complexity of a device but limiting the kinesthetic forces it can apply. However, previous research has shown that cutaneous stimulation alone is sufficient to render virtual objects of different masses, friction, and stiffness \cite{schorr2017fingertip}. Additionally, it has been shown that as the number of DoF of the device increases, users use less grip force and rate the haptic feedback as more realistic \cite{suchoski_2019}. As research on wearable devices for the fingertips is still nascent, the most desirable device characteristics, such as number of DoFs, that maximize the trade-off between realism and cost or size are unclear. It is an open question whether the improvement in realism is worth the additional cost and complexity of adding additional DoF.

Parallel kinematic mechanisms are mechanisms that can achieve large forces and high DoF with a small form factor. However, manufacturing techniques for creating wearable parallel haptic devices for the fingertip are limited and rely on rigid linkages and mechanisms, which are often expensive, difficult to manufacture at small sizes, and are not easily scalable for production. This is in contrast to origami (foldable) manufacturing techniques, which have recently been explored in haptic applications. Paik et al. demonstrated the utility of origami robotics in haptics in a 3-DoF origami force feedback device \cite{mintchev2019portable}. Their origami device is integrated into a holdable interface that rests in the palm of the user. They demonstrated how the device could be used for virtual reality, teleoperation, and surgical applications. However, this manufacturing technique has not been explored for devices with more DoF or for cutaneous or wearable haptic devices.

In this paper, we present the benefits of origami (foldable) manufacturing processes as an alternative for cutaneous fingertip devices. We show the design and fabrication of a 4-DoF parallel haptic device that uses origami fabrication methods and demonstrate that the haptic device is able to deliver normal, shear, and torsion feedback to a user. The manufacturing techniques enable the design of multi-DoF parallel mechanisms for haptics that have compact form factors, and affordable and scalable manufacturing processes. We demonstrate that origami manufacturing techniques can create a haptic device that can apply 2\,N of force and 5\,N$\cdot$mm of torque to a user. Additionally, we characterize the workspace and bandwidth of the device. Finally, we show the device's utility in two demonstrations.

\section{Design and Fabrication}

\subsection{4-DoF Parallel Mechanism}

\begin{table}[b]
\centering
\caption{4-DoF Wearable Haptic Device Technical Specifications}
\begin{tabular}{|p{3cm}|p{3cm}|}
\hline
Motor                    & Maxon DC Motor DCX06M EB SL 3V     \\ \hline
Gear Head                & Maxon GPX06 A 57:1                        \\ \hline
Encoder                  & Maxon ENX 6 OPT, 128 Counts, 2 Channel  \\ \hline
Dimensions {[}mm{]}      & $80 \times 80 \times 45$    \\ \hline
Link Lengths {[}mm{]}    & $L_i = 17.5$, $l_i = 15$, \\ & $d_1 = 12.15$, $h_1 = 4.97$, \\ & $d = 7.5$, $h = 17.5$ \\ & for $i = 1, 2, 3, 4$                 \\ \hline
Weight {[}g{]}           & 50 g                          \\ \hline
Max Normal Force {[}N{]} & 2                            \\ \hline
Max Shear Force {[}N{]}  & 1.5                             \\ \hline
Max Torque {[}N$\cdot$mm{]}     & 5                             \\
\hline                          
\end{tabular}
\label{table:devicespecs}
\end{table}

The 4-DoF origami device is composed of three separate components: the finger mount, the motor base, and the origami assembly, which is comprised of four legs and a traveling plate with a tactor (Fig. \ref{fig:origami_device}). The finger mount orients the user's fingerpad in the device's workspace and grounds the device to the back of the finger. The finger is secured using Velcro straps placed along the intermediate and proximal phalanges.

The finger mount, origami device, and motors are mounted to the motor base. The motors are 3\,V Maxon DC Motor DCX06M EB SL with a 57:1 gear head a 128-count optical encoder. The origami device weighs 10\,g and the motor base weights 35\,g. The entire device fits into a box of 80\,mm $\times$ 80\,mm $\times$ 45\,mm. The motors are controlled using a Sensoray 826 DAQ running a PID controller for each motor. The desired forces are computed and the motors are controlled with 1900\,Hz loop frequency.

The kinematic architecture of the origami device is inspired by a 4-DoF manipulator proposed by Pierrot et al. in \cite{pierrot2009optimal}. Each origami leg is composed of two rotational joints followed by a parallelogram (equivalent to a 1-DoF $\pi$ 4-bar linkage). The parallelogram on each joint provide constraints that confine the shear (x- and y- directions) movement in a plane orthogonal to the z-direction. An additional rotational joint attaches each of the four parallelograms to a corner of the traveling plate. The traveling plate contains another parallelogram, $C_{1\text{-}4}$, where one bar of the parallelogram connects to the tactor, which is the part of the device that interfaces with the user's skin. The tactor is the end effector of the robotic mechanism. The parallelogram on the traveling plate allows for rotation about the z-axis. The bottom of each leg attaches to a 4-bar linkage that is secured to a motor mounted to the bottom side of the motor base (Fig. \ref{fig:origami_device}A and B).

\subsection{Origami Fabrication Methods}

\begin{figure}[ht!]
    \includegraphics[width = \columnwidth]{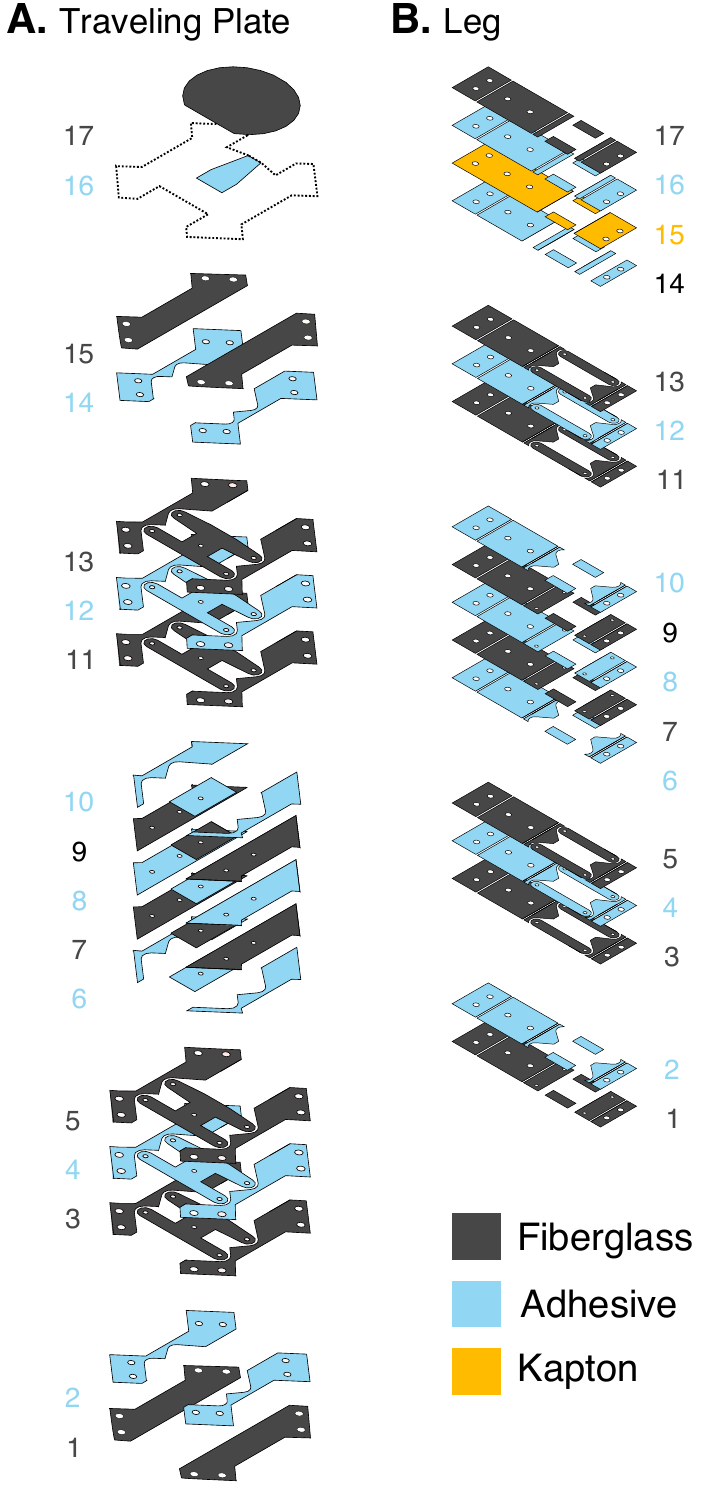}
    \caption{All layers for the traveling plate and one origami leg. (A) All 17 layers of the the traveling plate. Four 1\,mm pins were used in layers 3-13 to attach the 4-bar linkages to the remaining layers. The pins were added before adhering all layers in a heat press. Layer 17 interfaces directly with the skin of the fingertip. (B) All 17 layers of one origami leg. The Kapton layer (in yellow) is is the flexible layer that allows the two legs to rotate with respect to one another. The material that is removed during the release step is not shown in this image for ease of visualization.}
    \label{fig:fabrication}
\end{figure}

The device is manufactured using origami fabrication methods, wherein sheets of material are laser cut, combined in a heat press at 400\,psi and 400\,$^{\circ}$F, and laser cut again to release their final shape. Fiberglass sheets (0.005\,in), the same material used in circuit board manufacturing, acts as the rigid layers of the robotic structure. Kapton layers (13\,mm), a flexible material able to withstand high heat, are used to allow rotation at the joints. Dupont\textcopyright~Pyralux LF0100 25\,$\mu$m is a dry sheet adhesive that is used to bind together the layers of Kapton and fiberglass. We used the DPSS Lasers Inc. Samurai UV Marking System to laser cut the layers.

The origami fabrication method is used to construct the traveling plate, Fig. \ref{fig:fabrication}A, and four legs, Fig. \ref{fig:fabrication}B, of the origami assembly. All five parallelograms in the origami assembly use four 1mm dowel pins as pin joints in each of its four corners. The dowel pins go through layers 3-13 of both the traveling plate and legs in order to to constrain the parallelogram to the other layers. The traveling plate also connects to the tactor  Fig. \ref{fig:fabrication}A, Layer 17.

It is possible to laser cut and manufacture all four legs and the traveling together on one sheet so that only one release cut is necessary. However, we were limited by the size of our laser cutter, which was restricted to an 8\,cm x 8\,cm workspace. Consequently, the legs and tactor top were fastened using screws. 

\section{Kinematics and Control}
\begin{figure}[]
    \includegraphics[width = \columnwidth]{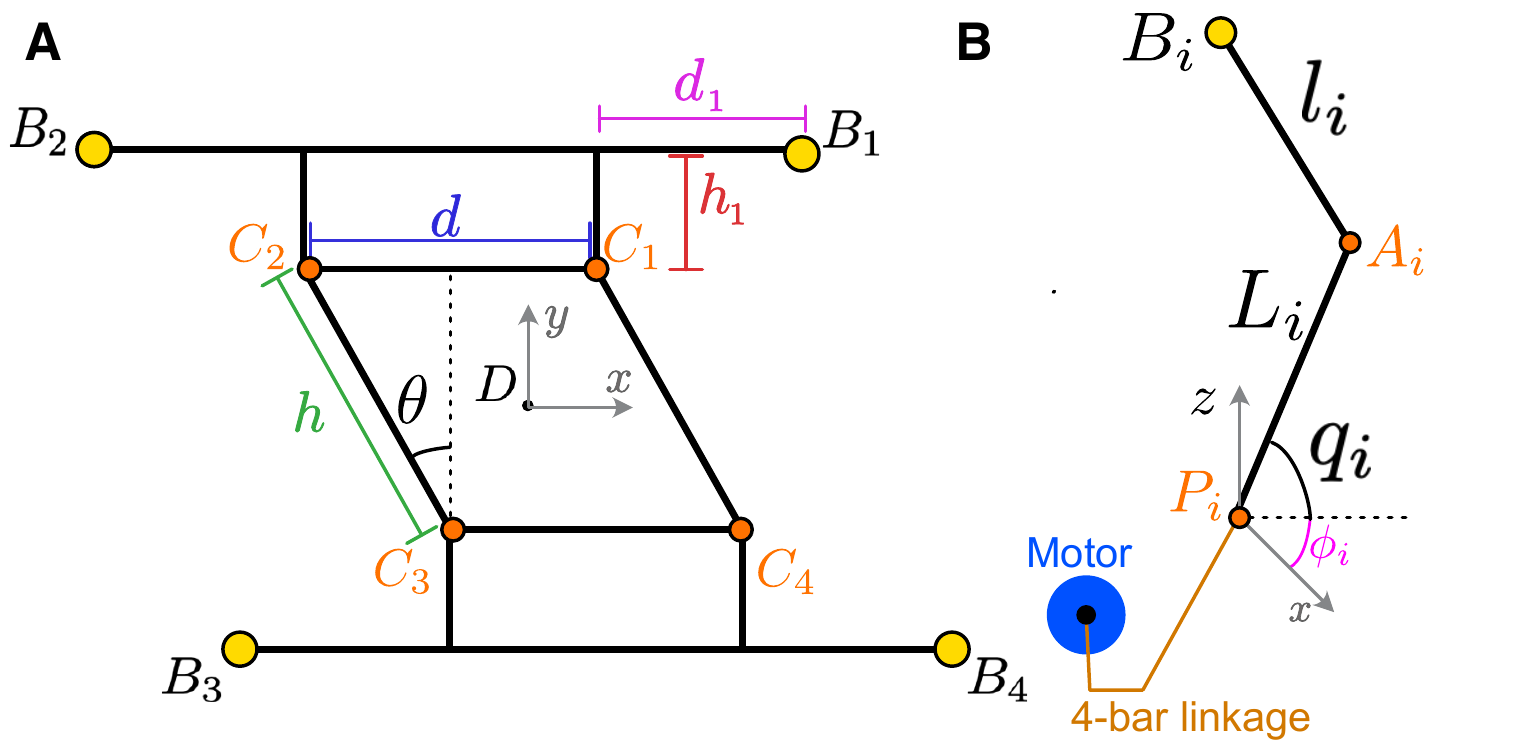}
    \caption{Diagram to describe relevant kinematic variables. (A) Kinematic variables of traveling plate. (B) Projection of leg in plane for ease of defining kinematic variables. $\vec{AB}_i$ is a parallelogram for all legs.}
    \label{fig:diagram}
\end{figure}

The forward kinematics of the 4-DoF parallel mechanism are presented by Pierrot et al. \cite{pierrot2009optimal}. We additionally found the solution to the inverse kinematics problem. First, we introduce the geometry and important parameters of the device, and then present the inverse and forward kinematics. 

\subsection{Kinematic Chain}
A schematic detailing the configuration of each leg, the traveling plate, and the kinematic variables is shown in Fig. \ref{fig:diagram}. The task space coordinates of the tactor position, which is defined as the center of the traveling plate, $D$ is designated by $x$, $y$, $z$, and $\theta$. The vector between the origin $O$ and $D$ is, $\overrightarrow{OD}=(x,y,z)^T$. Given the symmetry of the parallel robot, each of the 4 kinematic chains, $i=1,2,3,4$, has the same geometric representation and each can be analyzed independently. 

$A_i$, $B_i$, and $C_i$ are the centers of their respective joints and are pictured in Fig. \ref{fig:diagram}. We observe that the magnitude of the vector formed between $A_i$ and $B_i$ must be equal to the link length, $l_i$:

\begin{equation} \label{eq:leg_equality}
   \lvert\lvert \overrightarrow{A_i B_i} \rvert\rvert ^2 = l_i^2. 
\end{equation}
We define the center of each actuated joint, or the joints where the angle is directly controlled, to be $P_i=(x_i,y_i,z_i)^T$. We compute $\overrightarrow{A_i B_i}$ for each leg using the following relationship:
\begin{equation} \label{linkageSum}
    \overrightarrow{OP_i} + \overrightarrow{P_i A_i} + \overrightarrow{A_i B_i} + \overrightarrow{B_i C_i} = \overrightarrow{OD} + \overrightarrow{D C_i}.
\end{equation}
The vector $\overrightarrow{P_i A_i}$ is defined as:

\begin{equation}
    \overrightarrow{P_i A_i}=  \begin{bmatrix} L \cos \phi_i \cos q_i \\ L \sin \phi_i \cos q_i \\ -L \sin q_i \end{bmatrix},
\end{equation}
where $\phi_i$ is the angle of the linkage with respect to the origin. We define $\phi_1 = \frac{\pi}{4}$,  $\phi_2 = \frac{3\pi}{4}$,  $\phi_3 = \frac{5\pi}{4}$,  $\phi_4 = \frac{7\pi}{4}$, given the placement of the legs with respect to the axes. Additionally,  $\overrightarrow{OP_i}= P_i$. The vectors $\overrightarrow{DB_i}$  for our configuration are given in Appendix A.

Using Eq. \ref{eq:leg_equality}, we derive the following relationship between the joint angles $\vec{q} = (q_1, q_2, q_3, q_4)$ and the task space coordinates $x$, $y$, $z$, and $\theta$, where $\vec{DB}_i$ is dependent on $\theta$: 
\begin{equation}
    l_i^2= \left\lvert \left\lvert \begin{bmatrix} x \\ y \\ z\end{bmatrix} + \overrightarrow{DB_i}- \begin{bmatrix} L \cos \phi_i \cos q_i \\ L \sin \phi_i \cos q_i \\ -L \sin q_i \end{bmatrix} \right\rvert  \right\rvert^2.
    \label{eq:magnitudeEq}
\end{equation}


\subsection{Inverse Kinematics}
Solving Eq. \ref{eq:magnitudeEq} for $q_i$, $i=1,2,3,4$, provides us with the inverse kinematics of the system. For convenience we say that:
\begin{equation}
    \begin{bmatrix} X_i \\ Y_i \\ Z_i\end{bmatrix} = \begin{bmatrix} x \\ y \\ z\end{bmatrix} + \overrightarrow{DB_i}
    \label{simplified}
\end{equation}
We arrange Eq. \ref{eq:magnitudeEq} such that it has the form $I_i \sin{q_i} + J_i \cos{q_i} + K_i = 0$. We then find the joint angle $q_i$ to be:

\begin{equation}
    q_i = 2 \arctan \left( \frac{-I_i \pm \sqrt{\Delta}}{K_i - J_i} \right)
\end{equation}
where $\Delta_i = \sqrt{I_i^2 - K_i^2 + J_i^2}$. 
Given Eq. \ref{simplified}, we find the following values for $I_i$, $J_i$, and $K_i$
\[\begin{bmatrix} I_i \\ J_i \\ K_i \end{bmatrix} = \begin{bmatrix}2 Z_i L_i\\ - 2 X L_i\cos{\phi_i} - 2 Y_i L_i \sin{\phi_i} \\  L_i^2 - l_i^2 + X_i^2 + Y_i^2 + Z_i^2 \end{bmatrix}.\]

After computing $q_i$ for $i = 1, 2, 3,4$, the motor angle is determined by computing the corresponding angle of the 4-bar mechanism connecting the leg to the motor.


\subsection{Forward Kinematics}
 Nabat proposes solving the forward kinematics using an iterative method \cite{nabat2007robots, whitney1969resolved}. The iterative method was presented instead of the direct solution to the forward kinematics to because the direct solution is non-trivial.
 
 To use the iterative method, the current is updated using the Jacobian, $J$, as presented in \cite{pierrot2009optimal}, and the change in $\vec{q}$ over time. The current position of the end effector at time $t$ is defined as $\vec{p}_{t} = [x_{t}, y_{t}, z_{t}, \theta_{t}]^T$. The end effector position at the previous time step, $t-\Delta t$, is defined as $\vec{p}_{t-1} = [x_{t-\Delta t}, y_{t-\Delta t}, z_{t-\Delta t}, \theta_{t-\Delta t}]^T$. We define the change in the actuator angles between time steps as, $\Dot{\vec{q}} = (\vec{q}_{t} - \vec{q}_{ t-\Delta t})/\Delta t$, where $q_\tau = [q_{\tau,1}, q_{\tau,2}, q_{\tau,3}, q_{\tau,4}]^T$ and $q_{\tau,i}$ is the joint angle at time $\tau$ for leg $i$. The current position $\vec{p}_{t}$ can be calculated given an previous position $\vec{p}_{t-1}$ using the following update rule: 
\begin{equation}
    \vec{p_{t}} = \vec{p}_{t-\Delta t} + J \Dot{\vec{q}}
\end{equation} 

Alternatively, the forward kinematics can be solved directly by finding $x$, $y$, $z$, and $\theta$ in Eq. \ref{eq:leg_equality}. In this paper, we present the forward kinematics solution. 

We find the solution by subtracting Eq. \ref{eq:magnitudeEq} when $i = 1$ from each of Eq. \ref{eq:magnitudeEq} where $i = 2, 3, 4$. The resulting three equations are linear and can be written as a system of linear equations. Next, Cramer's rule is applied to solve for $x$, $y$, and $z$. The resulting equations describe the position of the end effector,  $x$, $y$, and $z$.

\begin{equation}
    x =\frac{ a_x  + b_x \sin{\theta} + c_x \cos^2{\theta} + \cos{\theta}(d_x + e_x \sin{\theta})} {2(a_d + b_d \cos{\theta} + c_d \sin{\theta})}
\end{equation}

\begin{equation}
    y =\frac{ a_y - b_y \sin{\theta} + c_y \sin^2{\theta} + \cos{\theta}(d_y + e_y \sin{\theta})} {2(a_d + b_d \cos{\theta} + c_d \sin{\theta})}
\end{equation}

\begin{equation}
    z =\frac{ a_z + b_z \cos{\theta} + c_z \cos{\theta}^2 + \sin{\theta}(d_z + e_z \cos{\theta}) +f_z \sin^2{\theta}} {2(a_d + b_d \cos{\theta} + c_d \sin{\theta})}
\end{equation}

All $a_n$, $b_n$, $c_n$, $d_n$, $e_n$, and $f_n$ where $n \in \{x, y, z, d\}$ are coefficients whose values are included in Appendix B. The resulting values of $x$, $y$, and $z$ are substituted back into the Eq. \ref{eq:magnitudeEq} when $i = 1$, to solve for $\theta$. 

The resulting polynomial is an 8th order polynomial with coefficients $c_i$, $i = 0~ ... ~8$. We define $t = \tan{\frac{\theta}{2}}$, such that the polynomial has the following form, 
\begin{equation} \label{eq:thetapoly}
c_0 + c_1 t + c_2 t^2 + c_3 t^3 + c_4 t^4 + c_5 t^5 + c_6 t^6 + c_7 t^7 + c_8 t^8  = 0. 
\end{equation}
The coefficients $c_i$, $i = 0~ ... ~8$ are presented in Appendix B. The value of $t$ and consequently $\theta$ can be found using numerical methods, such as Newton's root-finding algorithm. The direct method of calculating the inverse kinematics can be used when the iterative method is not sufficient or when the initial actuator angles, and not the the initial end-effector position, are known.

\section{Experiments and Results}
\subsection{Bandwidth Testing}

\begin{figure}[!t]
\includegraphics[width=\columnwidth]{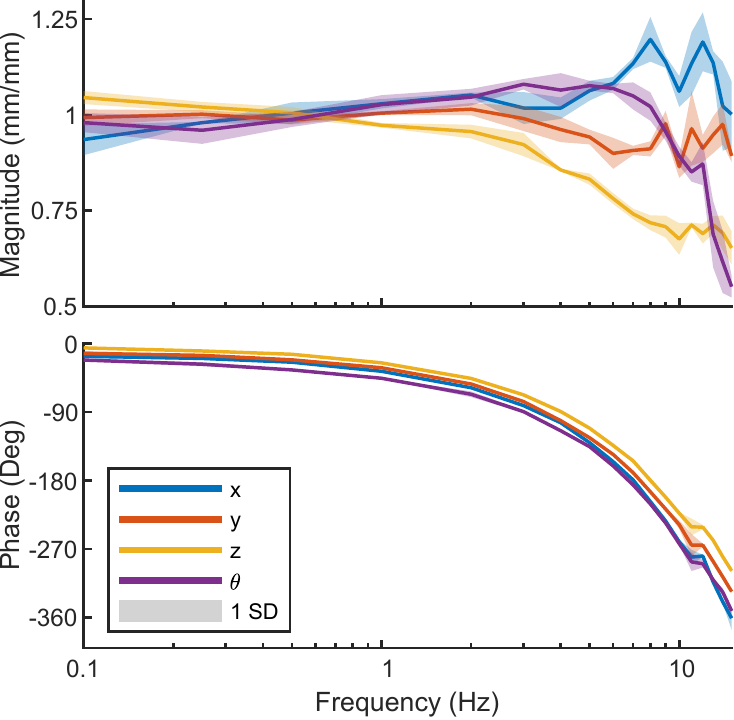}
    \caption{Bode plot for each of the device DoFs up to 15\,Hz. All DoFs have different cutoff frequency and the overall bandwidth of the device is limited by the z-direction, which has a -3dB cutoff of 9Hz.}
    \label{fig:bw_plot}
\end{figure}

Bandwidth testing was performed in each of the four degrees-of-freedom. The position and orientation of the device end-effector was measured using the MicronTracker (ClaroNav, Toronto, ON, Canada), a marker-based optical tracking system. 

The position and orientation of the end-effector were measured relative to the base of the device. The z-axis was defined as normal to the plane formed by the motor base, and the x-axis defined such that x-direction movements commanded by the device drivers were approximately parallel. This alignment was achieved by projecting the first principle direction obtained from the singular value decomposition of data points captured from commanded x-direction movements into the base plane and defining the direction of that vector as the base frame x-axis.

For angular displacements about the z-axis, the rotations were measured directly with respect to the MicronTracker and were projected about the base frame z-axis using the Swing-Twist Decomposition \cite{dobrowolski2015swingtwist}, with the base frame z-axis being defined as the twist axis. 

2\,mm amplitude sinusoidal oscillations were commanded in the x, y, and z-directions while 14$^\circ$ amplitude sinusoidal oscillations were commanded in the $\theta$-direction. Three frequency sweeps were conducted, and the magnitudes and phase lags at each frequency were averaged and used to generate the Bode plot shown in Fig.\,\ref{fig:bw_plot}. Data points were collected at 0.1, 0.25, 0.5\,Hz, and from 1 to 15\,Hz in 1\,Hz increments.

Each of the DoFs has a different cutoff frequency, and the overall bandwidth of the device is limited by the z-direction, which has a -3\,dB cutoff of 9\,Hz. We observed multiple peaks in the amplitude response for all directions and the phase response indicates the system has multiple poles. These results could be due to non-linear behavior induced by friction between layers at the pin joints in the parallelograms on each leg and on the tactor, or by the inherent flexibility of the origami assembly.

\subsection{Device Workspace}
\begin{figure}
    \centering
    \includegraphics[width = \columnwidth]{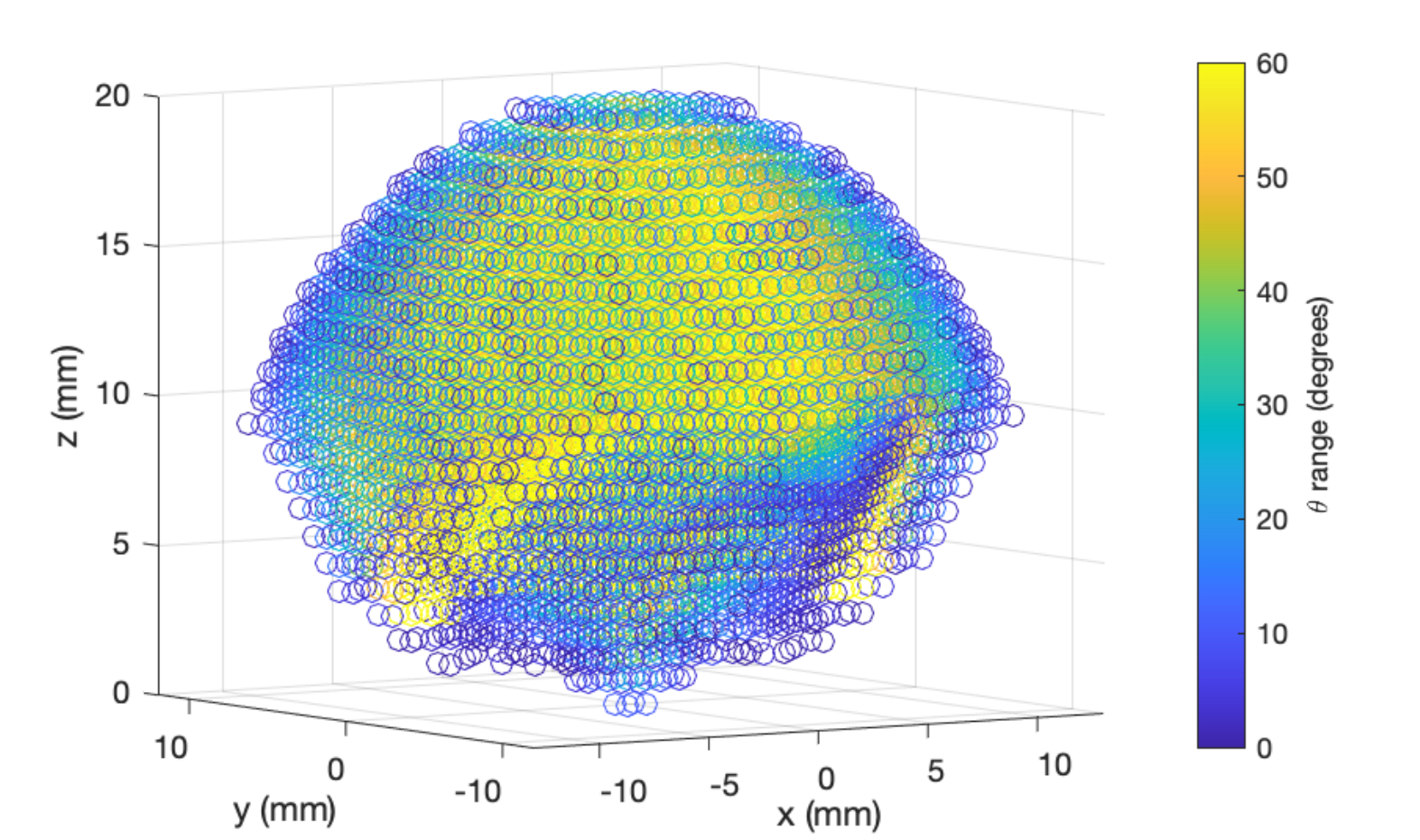}
    \caption{3D workspace of 4-DoF origami device. The kinematics model shows that the workspace spans $\pm$ 24\,mm in the x-direction, $\pm$ 26\,mm in the y-direction, $\pm$ 18\,mm in the z-direction, and $\pm 30 ^{\circ}$ in the $\theta$-direction.}
    \label{fig:workspace}
\end{figure}

Given the motor torques, the workspace of the device can be determined in simulation. The maximum movement in the $x$, $y$, $z$, and $\theta$-directions is $\pm 13$, $\pm 12$, $\pm 9$ mm, and $\pm 30^{\circ}$, respectively. However, when the workspace is defined as a cube where every position can achieve minimum $60^{\circ}$ total rotation, the reachable workspace is confined to an $8~\mathrm{mm} \times 10~\mathrm{mm} \times 8~\mathrm{mm}$ cube. Fig. \ref{fig:workspace}A shows the device workspace without constraints. The size of the workspace can be increased by increasing the length of the legs. Consequently, the size of the device and workspace can be modified to accommodate a desired application.

\section{Force Measurements}

The maximum forces in all directions (Table \ref{measuredforces}) are measured using an ATI Nano-17 force/torque sensor (ATI Industrial Automation, Apex, NC, USA). The force/torque sensor and the base of the 4-DoF constrained device are mounted so that they cannot move relative to one another. An alternative traveling plate is used to connect the force/torque sensor securely to the legs of the 4-DoF constrained device. Four repetitions of force measurements are collected in each DoF and the mean and standard deviations are reported in Table \ref{measuredforces}.

The results show that the device can produce $\pm$1-2 N of force in each direction and greater than $\pm$5 (N$\cdot$mm) of torque.  We observe that the y-direction produces a larger force than the x-direction. This is expected as the traveling plate is a rectangle, not a square, which results a different moment arm for different legs. We also observe that the positive y-direction forces and $\theta$ torques are smaller than the negative y-direction forces and $\theta$ torques. This could be due to manufacturing inaccuracies.

\begin{table}
\centering
\caption[Measured forces.]{Mean and standard deviation of forces in all DoFs.}
\begin{tabular}{|c|c|c|}
\hline 
direction            & mean & std \\\hline 
x+ (N)               & 1.05 & 0.09               \\\hline 
x- (N)               & 1.06 & 0.05               \\\hline 
y+ (N)               & 1.25 & 0.08               \\\hline 
y- (N)               & 1.61 & 0.08               \\\hline 
z+ (N)               & 1.39 & 0.13               \\\hline 
$\theta$+ (N$\cdot$ mm) & 5.6 & 1.4               \\\hline 
$\theta$- (N$\cdot$ mm) & -8.2 & 0.8               \\\hline 
\end{tabular}
\label{measuredforces}
\end{table}

\begin{figure*}[]
    \centering
    \includegraphics[width =0.9\textwidth]{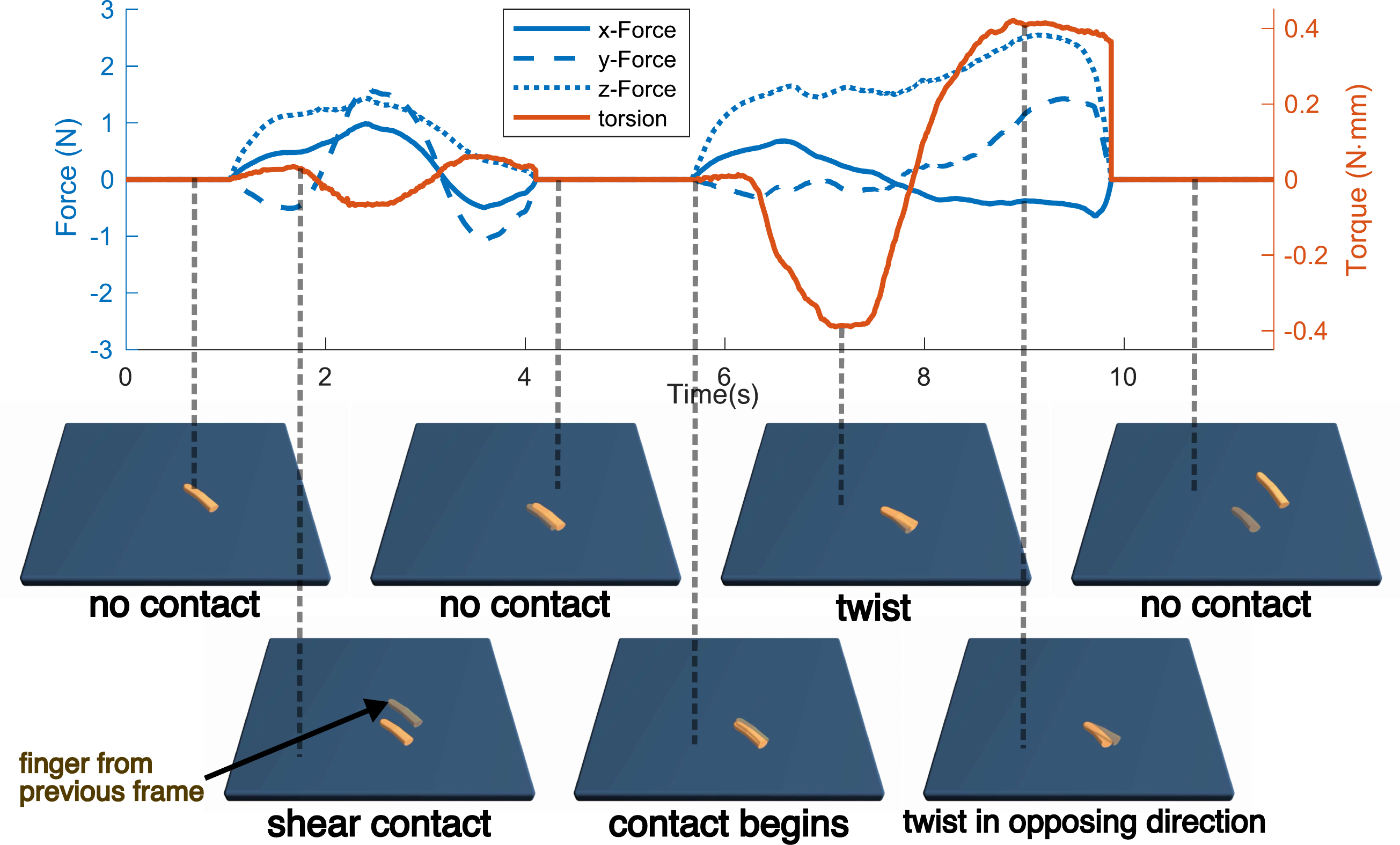}
    \caption{Demonstration of shear and torsion feedback. The top figure shows the desired force (in N) and torque (in N$\cdot$mm) throughout time. From 1\,s-4\,s, the user is creating shear feedback with the object's surface. From 6\,s-10\,s, the user is rotating their finger, creating torsion feedback. The surface (in blue) is grounded and cannot move within the virtual environment. The seven images show snapshots of the finger's location in the virtual world.  The dashed grey line shows the correspondence between the image and the forces rendered by the device. The finger from the proceeding image is shown (see-through and lighter in color) in each current image to provide a reference for how the finger movement is changing. }
    \label{fig:demo_interaction}
\end{figure*}

\begin{figure*}
    \centering
    \includegraphics[width =0.9\textwidth]{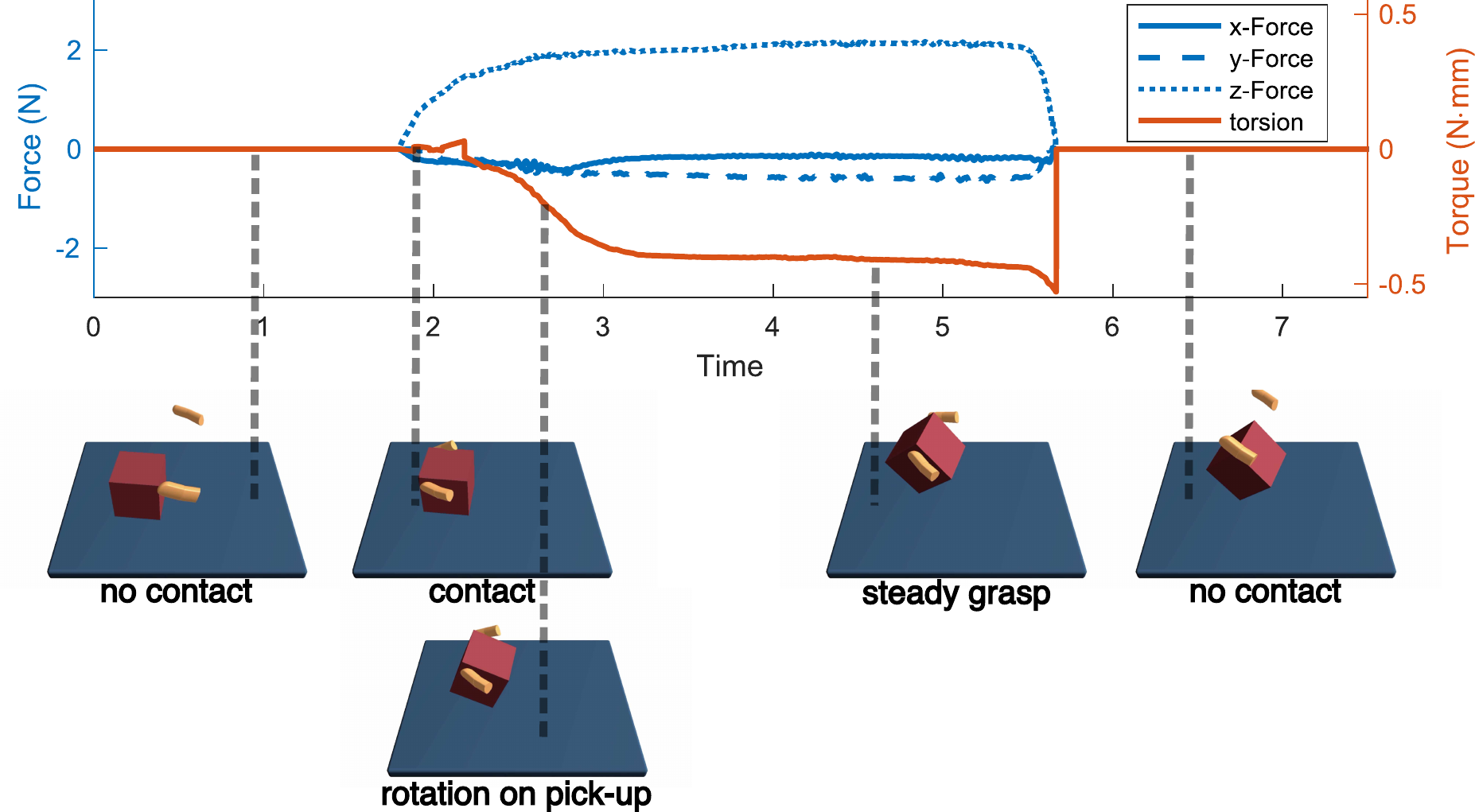}
    \caption{Demonstration of user picking up and releasing block. The top figure shows the desired force (in N) and torque (in N$\cdot$mm) throughout time. The surface (in blue) is grounded and cannot move within the virtual environment. The five images are snapshots of the user picking up and releasing a block in the virtual world. When the user initiates contact, the normal force (z-direction) increases. As the user picks-up the block, it rotates in their fingertips increasing the torsion applied to the fingers. When the block is steady in the user's fingertips, a constant normal force, shear force in the negative y-direction, and twist are applied to the user until the block is released. The dashed grey line shows the correspondence between the image and the forces rendered by the device. }
    \label{fig:demo_pick}
\end{figure*}

\section{VR Device Demonstration}

We show the effectiveness of the device in two demonstrations where the device is mounted to the index finger of the right hand of the user. The CHAI3D framework is used as the rendering environment \cite{Conti03}. CHAI3D uses the god-object algorithm \cite{zilles1994constraint} to compute the interaction forces and the friction-cone algorithm to compute the frictional forces  \cite{harwin2002improved}. CHAI3D does not include torsion computations for the fingerpad, so we developed a module to compute the torsion shear forces using the soft finger-proxy algorithm (the slip condition was not included) \cite{barbagli2004simulating}. Software limits were set to confine the workspace to $7~\mathrm{mm} \times 9~\mathrm{mm} \times 7~\mathrm{mm}$. The user's index finger position is detected using a magnetic tracker (3D Guidance trakSTAR, NDI, Waterloo, ON, Canada). During the demonstration, the user sees an avatar of a finger that represents the index finger's location in the virtual environment.

 In the first demonstration, Fig. \ref{fig:demo_interaction}, the user places their finger on the surface of an object, resulting in an increase in the  z-forces. In both demonstrations, the blue object is grounded, so it cannot move in the virtual world. The blue object is programmed with a linear stiffness of $500\,\mathrm{N}/\mathrm{m}$, static friction of $2.0\,\mathrm{N}/\mathrm{m}$, and dynamic friction of $1.8\,\mathrm{N}/\mathrm{m}$. After making contact with the object the user moves their finger back and forth primarily in the y-direction. We observe that there is some torsion displayed to the finger but the haptic feedback is principally shear feedback. After breaking and making contact with the table once again, the user twists their finger clockwise and then counter-clockwise, primarily activating the torsion degree of freedom.

In the second demonstration, the user picks up the object using their index finger and thumb, which are each tracked using a magnetic tracker and rendered using an avatar. The thumb receives no haptic feedback. The desired feedback for the fingertip device during the pick-and-release task is showed in Fig.\,\ref{fig:demo_pick}. The $z$-force increases when contact is initiated around 2\,s. As the user starts to pick up the block, it begins to rotate in the user's fingers, resulting in a torsion applied to the fingertips. When the block stops rotating, the torsion reaches equilibrium, until the cube is released. 


\section{Discussion and Conclusions}

Little research has been conducted to understand how torque influences human perception and performance during manipulation tasks with wearable cutaneous devices for the fingertip. However, the origami parallel mechanism and fabrication techniques presented in this paper introduce design principles that can be used to create devices that can investigate fundamental questions about rotational cues for the fingertip. In future work, we plan to investigate user performance as we vary which DoFs are used during manipulation tasks to determine the DoFs that are the most critical for user performance. Additionally, we will investigate how other parallel kinematic structures, with up to 6 DoFs, can be created using origami fabrication methods.

Although the current origami device is able to rotate with with a maximum angle of $\pm 30^{\circ}$, it is possible that additional rotational displacement may be useful for even more compelling feedback. Researchers have shown that the minimum orientation-change threshold for shear feedback is 14-34$^{\circ}$ and the number increases to 64$^{\circ}$ when the individual is engaging in active movement, such as movement that occurs during manipulation tasks \cite{vitello2006instance}. Although these experiments do not directly measure torques or rotational displacements, they indicate that future devices may require larger displacements for users to be able to feel a large range of differential torsional cues. This can be achieved using the device presented in this paper by amplifying the angular displacement using a belt and pulley, as seen in \cite{pierrot2009optimal}, or other mechanisms.

In \cite{young2019implementation}, the researchers found that that rotation cues were perceived with the highest accuracy when stimulating the center of the fingerpad. However, the traveling plates of the current device are constrained to remain flat and the fingermount cannot be adapted to users with different finger curvature or size. Without accurate placement of the device in the center of the finger, the user might have diminished torsion perception. Creating an adjustable tactor or fingermount may improve the overall performance of the device across users.

\addtolength{\textheight}{-12cm}   



\section*{APPENDIX A}

The vector $\overrightarrow{DB_i}$  varies for each link.

\[\overrightarrow{DB_1} =  \begin{bmatrix} -\frac{1}{2}h\sin\theta+d_1+\frac{d}{2} \\ \frac{1}{2}h\cos\theta+h_1 \\ 0 \end{bmatrix} ,~ \overrightarrow{DB_2} =  \begin{bmatrix} -\frac{1}{2}h\sin\theta-d_1-\frac{d}{2} \\ \frac{1}{2}h\cos\theta+h_1 \\ 0 \end{bmatrix}\]

\[\overrightarrow{DB_3} =  \begin{bmatrix} \frac{1}{2}h\sin\theta-d_1-\frac{d}{2} \\ -\frac{1}{2}h\cos\theta-h_1 \\ 0 \end{bmatrix},~\overrightarrow{DB_4} =  \begin{bmatrix} \frac{1}{2}h\sin\theta+d_1+\frac{d}{2} \\ -\frac{1}{2}h\cos\theta-h_1 \\ 0 \end{bmatrix}\]

\section{APPENDIX B}

The coefficients for the inverse kinematics are composed of long equations that are not intuitive for the reader. Consequently, we have created Wolfram Mathematica and MATLAB files of the inverse kinematics, which are available at \url{https://github.com/sophiarw/4-DoFForwardKinematics} .

\section*{ACKNOWLEDGMENT}

The authors thank Bernie Roth and Dylan Black for their mathematical insights and expertise.


\bibliographystyle{IEEEtran}
\bibliography{file.bib}

\end{document}